\documentclass[11pt,a4paper]{article}
\usepackage{times,latexsym}
\usepackage{url}
\usepackage[T1]{fontenc}
\usepackage{diagbox}
\usepackage[acceptedWithA]{tacl2018v2}

\newtheorem{definition}{Definition}
\newcommand{\commentout}[1]{}

\usepackage[]{tacl2018v2}
\usepackage{amsmath}
\usepackage{subcaption}
\usepackage{multirow}
\usepackage{algorithm}
\usepackage{algorithmic}
\usepackage{ifthen}
\usepackage{color}
\usepackage[utf8]{inputenc}
\usepackage{url}
\usepackage{multicol}
\usepackage{tikz}
\usepackage{booktabs}
\usepackage{csquotes}
\usepackage{soul}
\usepackage{float}
\usetikzlibrary{fit}
\usetikzlibrary{patterns}
\usepackage{pifont}

\usepackage{enumitem}
\usepackage{tabularx}

\DeclareGraphicsExtensions{.png,.pdf}

\colorlet{centity1}{red!20}
\colorlet{centity2}{orange!20}
\colorlet{centity3}{orange!20}
\colorlet{canswer}{blue!20}
\colorlet{csent}{gray!20}

\newcommand{\espan}[3][yellow]{%
\sethlcolor{#1}
  {\hl{#2}}$_{#3}$
}

\newcommand{\sspan}[2][yellow]{%
\sethlcolor{#1}
  {\hl{#2}}
}

\newcommand{\spanone}[2]{\espan[centity1]{#1}{#2}}
\newcommand{\spantwo}[2]{\espan[centity2]{#1}{#2}}

\newcommand{\spanans}[2]{\espan[canswer]{#1}{#2}}
\newcommand{\spansent}[1]{\sspan[csent]{#1}}

\title{QED: A Framework and Dataset for Explanations in Question Answering}

\author{
Matthew Lamm$^1$\thanks{$\;$Work done during internship at Google.}$\,$, Jennimaria Palomaki$^2$, Chris Alberti$^2$,\\
{\bf 
Daniel Andor$^2$, Eunsol Choi$^3$\thanks{$\;$Work done at Google.}$\,$, Livio Baldini Soares$^2$, Michael Collins$^2$}\\
\\
$^1$Department of Linguistics, Stanford University\\
$^2$Google Research \\
$^3$Department of Computer Science, The University of Texas at Austin\\
  {\small {\sf mlamm@stanford.edu, \{jpalomaki, chrisalberti, danielandor, liviobs, mjcollins\}@google.com, eunsol@cs.utexas.edu}} 
}

\ifthenelse{\isundefined{\definition}}{\newtheorem{definition}{Definition}}{}
\ifthenelse{\isundefined{\assumption}}{\newtheorem{assumption}{Assumption}}{}
\ifthenelse{\isundefined{\hypothesis}}{\newtheorem{hypothesis}{Hypothesis}}{}
\ifthenelse{\isundefined{\proposition}}{\newtheorem{proposition}{Proposition}}{}
\ifthenelse{\isundefined{\theorem}}{\newtheorem{theorem}{Theorem}}{}
\ifthenelse{\isundefined{\lemma}}{\newtheorem{lemma}{Lemma}}{}
\ifthenelse{\isundefined{\corollary}}{\newtheorem{corollary}{Corollary}}{}
\ifthenelse{\isundefined{\alg}}{\newtheorem{alg}{Algorithm}}{}
\ifthenelse{\isundefined{\example}}{\newtheorem{example}{Example}}{}
\newcommand\entails{\mkern2mu\vdash} 

\newcommand\ml[1]{\textcolor{red}{[ML:\@ #1]}}
\newcommand\ca[1]{\textcolor{red}{[CA:\@ #1]}}
\newcommand\mc[1]{\textcolor{blue}{[MC:\@ #1]}}
\newcommand\jp[1]{\textcolor{red}{[JP:\@ #1]}}
\newcommand\ec[1]{\textcolor{olive}{[EC:\@ #1]}}
\newcommand\lbs[1]{\textcolor{green}{[LBS:\@ #1]}}
\newcommand\da[1]{\textcolor{green}{[DA:\@ #1]}}

\renewcommand\ml[1]{}
\renewcommand\ca[1]{}
\renewcommand\mc[1]{}
\renewcommand\jp[1]{}
\renewcommand\ec[1]{}
\renewcommand\lbs[1]{}
\renewcommand\da[1]{}

\begin{document}
\maketitle
\begin{abstract}
A question answering system that in addition to providing an answer provides an {\em explanation} of the reasoning that leads to that answer has potential advantages in terms of debuggability, extensibility and trust.
To this end, we propose QED, a linguistically informed, extensible framework for explanations in question answering. 
A QED explanation specifies the relationship between a question and answer according to formal semantic notions such as referential equality, sentencehood, and entailment. 
We describe and publicly release an expert-annotated dataset of QED explanations built upon a subset of the Google Natural Questions dataset, and report baseline models on two tasks -- post-hoc explanation generation given an answer, and joint question answering and explanation generation. In the joint setting, a promising result suggests that training on a relatively small amount of QED data can improve question answering.
In addition to describing the formal, language-theoretic motivations for the QED approach, we describe a large user study showing that the presence of QED explanations significantly improves the ability of untrained raters to spot errors made by a strong neural QA baseline.
\end{abstract}

\section{Introduction}
\label{sec:intro-3}

\begin{figure}[th!]
\centering
\input{intro-figure.tex}
\caption{QED explanations decompose the question-passage relationship in terms of referential equality and predicate entailment.\label{fig:intro-example}}
\vspace{-2ex}
\end{figure}

Question Answering (QA) systems can enable efficient access to the vast amount of information that exists as text~\cite[i.a.]{Rajpurkar_2016, 47761, clark2019boolq, Reddy_2019}. Modern neural systems have made tremendous progress in QA accuracy in recent years~\cite{Devlin2019BERTPO}. However, they generally give no explanation or justification of how they arrive at an answer to a question.  Models that in addition to providing an answer can explain their reasoning may have significant benefits pertaining to trust and debuggability~\cite{doshi2017towards, ehsan2019automated}.

Critical questions then, are what constitutes an {\em explanation} in question answering, and how can we enable models to provide such explanations. In an effort to make progress on these questions, in this paper we make the following contributions: (1) we introduce QED\footnote{QED stands for the Latin ``quod erat demonstrandum'' or ``that which was to be shown''.}, a linguistically grounded definition of QA explanations; and (2) we describe a corpus of QED annotations based on the Natural Questions~\cite{47761}. The QED corpus has been released publicly.\footnote{\url{https://github.com/google-research-datasets/QED}.} 

Figure~\ref{fig:intro-example} shows a QED example. Given a question and a passage, QED represents an explanation as a combination of discrete, human-interpretable steps: (1) identification of a sentence implying an answer to the question, (2) identification of noun phrases in both the question and answering sentence that refer to the same thing, and (3) confirmation that the predicate in the sentence entails the predicate in the question once referential equalities are abstracted away. 

This choice of explanation makes use of core semantic relations---referential equality and entailment---and thus has well-understood formal properties. (See Section~\ref{sec:motivation} for further discussion.) In addition, we found that this way of decomposing explanations has high coverage (77\% on the Natural Questions corpus\footnote{Instances with annotated short answers, omitting table passages.}). Since QED decomposes the QA process into distinct subproblems, we also believe that it should enable research directions aimed at extending or improving upon extant QA systems.

In what follows, after contextualizing the present work in the broader discussion on explainability, we present a formal definition of QED explanations. We then describe the dataset of QED annotations (7638/1353 train/dev examples), including discussion of the distribution of linguistic phenomena exhibited in the data. We move to propose four potential tasks, of varying complexity, related to the QED framework, and use the QED annotations to train and evaluate baseline models on two of these. Additionally, we describe a rater study which shows how the presence of QED explanations can help users identify errors made by an automated QA system.

\section{Motivation: The Need for Explanations in Question Answering}
\label{sec:motivation}
We take as our departure point the following passage from~\citet{ehsan2018rationalization} concerning explainable AI:

\begin{quote}
    {\em \commentout{Explainable AI refers to artificial intelligence and machine learning techniques that can provide human understandable justification for their behavior.}
    Explainability is important in situations where human operators work alongside autonomous and semi-autonomous systems because it can help build rapport, confidence, and understanding between the agent and its operator. In the event that an autonomous system fails to complete a task or completes it in an unexpected way, explanations help the human collaborator understand the circumstances that led to the behavior, which also allows the operator to make an informed decision on how to address the behavior.}
\end{quote}

This quote refers to AI and ML systems in general, but is highly relevant to QA systems. Explanations can help users understand and trust a QA system, and can help them to work in tandem with a QA system to fulfill their information needs. Explanations can also help system builders to understand and debug QA systems, and also to extend them.

QED makes a particular choice about the form of explanations for QA. In particular, it decomposes the question-answer relationship according to known semantic and syntactic categories -- sentence, reference (and referential equality), predicate, and entailment. The explanations provided in QED are discrete structured objects, as opposed, for example, to  ``heat map"-style explanations (attention distributions, or other real-valued, word-level feature importance measures)~\cite{jacovi2020towards}. 

One major goal in developing QED is to define models which provide {\em faithful} explanations; that is, explanations that in some sense truly reflect the underlying computation or reasoning performed by a question-answering model. (See Section~\ref{sec:discuss} for more discussion.) Another major goal, which is closely related to faithfulness, is to develop models that have a sound basis in concepts from cognitive science and linguistics, and are thus closer to human reasoning. For example reference, a core component of QED, is fundamental to semantics and cognition~\cite{russell1905denoting,clark1981definite,tomasello2007new}.

\commentout{It is an open question as to what constitutes a good explanation~\cite{lipton2001good}, and the answer depends on the task at hand. A major inflection point in the discussion is the notion of \textit{faithfulness}~\cite{ross2017right, lipton2016mythos}. We say a model's explanations are faithful when there is a causal relationship between an explanation and a prediction. That is, when  an explanation changes, the outputs change accordingly. When this is not true, we say a model generates \textit{rationales}, which have the appearance of justifying its outputs, but without causal guarantees~\cite{ehsan2018rationalization}.

QED is defined with faithful QA in mind, and is committed to the cognitive reality of reference (see, e.g.~\cite{russell1905denoting,clark1981definite,partee1986noun,tomasello2007new}) and entailment. We can say, definitively, that in order for a sentence to answer a question about a thing, its meaning must involve that thing in a very particular sense. Posed counterfactually, when you break referential equality, you break answerhood, and the same argument follows for predicate entailment. It seems, then, that unlike other intelligent behavior that may permit of post-hoc rationalization at best \cite{ehsan2019automated}, QA is a form of high-level linguistic reasoning that is in fact amenable to strong explanation.}

\section{Annotation Definition}

\newcommand{\nq}{m}
\newcommand{\nc}{n}
\newcommand{\eq}{e}
\newcommand{\ans}{a}
\newcommand{\sent}{s}
\newcommand{\senti}{{\sent_0}}
\newcommand{\sentj}{{\sent_1}}
\newcommand{\phrases}{{\cal S}}
\newcommand{\qphrases}{{\cal Q}}
\newcommand{\cphrases}{{\cal C}}

We now describe the form of QED annotations. Section~\ref{sec:approachoverview}  gives an overview of the annotation process. Section~\ref{sec:approachformal} then gives a formal definition, which is extended in Section~\ref{sec:bridge}.

\subsection{An Overview of the Approach}

\label{sec:approachoverview}

\commentout{
\begin{figure}

\noindent
{\bf Question:} how many seats in university of michigan stadium

$\;$

\noindent
{\bf Passage:} Michigan Stadium, nicknamed “The Big House”, is the football stadium for the University of Michigan in Ann Arbor, Michigan. It is the largest stadium in the United States, the second largest stadium in the world and the 34th largest sports venue. Its official capacity is 107,601.

$\;$

\noindent
{\bf Explanation:}

\noindent
{\bf Sentence:}
Its official capacity is 107,601.

\noindent
{\bf Referential equality annotations/Answer:}

\noindent
how many seats in [=1 university of michigan stadium]

\noindent
[=1 Its] official capacity is [=A 107,601].

\caption{(a) An example question/passage pair used as input to the annotation process. (b) The explanation for this example.}
\end{figure}}

We will use the following example to illustrate the approach:

$\;$

\noindent
\framebox{\parbox{2.9in}{
\begin{normalsize}
\noindent
{\bf Question:} how many seats in university of michigan stadium

\noindent
{\bf Passage:} Michigan Stadium, nicknamed “The Big House”, is the football stadium for the University of Michigan in Ann Arbor, Michigan. It is the largest stadium in the United States and the second largest stadium in the world. Its official capacity is 107,601.
\end{normalsize}}}

The annotator is presented with a question/passage pair. Annotation then proceeds in the following four steps:

\paragraph{(1) Single Sentence Selection.} The annotator identifies a single sentence in the passage that entails an answer to the question assuming that coreference and bridging anaphora (see Section~\ref{sec:bridge}) have been resolved in the sentence.\footnote{If it is not possible to find a sentence that satisfies these properties---typically because the answer requires inference beyond coreference/bridging that involves multiple sentences---the annotator marks the example as not  possible. See Section~\ref{sec:annotations}.}

In the above example, the following sentence entails an answer to the question, and would be selected by the annotator:

\vspace{-1ex}
\begin{quote}
{\em Its official capacity is 107,601.}
\end{quote}
\vspace{-1ex}

    This follows because given the passage context, ``Its'' refers to the same thing as the NP ``university of michigan stadium'' in the question, and the predicate in the sentence, ``X's official capacity is 107,601'', entails the predicate in the question ``how many seats in X''.
    
\paragraph{(2) Answer Selection.} The annotator highlights a short answer span (or spans) in the answer sentence. In the above example the annotator would mark the following (answer shown with [=A...]):
    
\begin{quote}
    {\em Its official capacity is [=A 107,601].}
\end{quote}
    
In addition if the answer appears in the sentence in the form of a pronoun, bridged reference or underspecified NP, the annotator resolves the underlying coreference within the passage (see Section~\ref{sec:bridge} for more discussion).

\paragraph{(3) Identification of Question-Sentence Noun Phrase Equalities.} The annotator marks referentially equivalent noun phrases, or noun phrases that refer to the same thing, in the question and the answer sentence. This includes reference not only to individuals and other proper nouns, but also to generic concepts.

In our example the annotator would mark the following two noun-phrases (marked with the [=1~...] annotations) as referentially equivalent:

\begin{small}
\begin{tabbing}
{\em how many seats in [=1 university of michigan stadium]}\\
{\em [=1 Its] official capacity is [=A 107,601]}
\end{tabbing}
\end{small}

\paragraph{(4) Extraction of an Entailment Pattern.} As a final, automatic step, an entailment pattern can be extracted from the annotated example by abstracting over referentially equivalent noun phrases, and the answer. In the above example the entailment pattern would be as follows:

\begin{quote}
{\em how many seats in X}

\noindent
{\em X's official capacity is ANSWER}
\end{quote}

\subsection{A Formal Definition}

\label{sec:approachformal}

\commentout{We now give a formal definition of QED annotations.\ml{redundant given subsection title}} 
An annotator is presented with a question $q$ that consists of $\nq$ tokens $q_1 \ldots q_{\nq}$, along with a passage $c$ consisting of $\nc$ tokens $c_1 \ldots c_{\nc}$.

The QED annotation is  a triple $\langle \sent, \eq, \ans \rangle$ where:

\begin{itemize}
    \item $\sent$ is a sentence within the context $c$. Specifically $\sent$ is a pair $\senti, \sentj$ indicating that the sentence spans words $c_{\senti} \ldots c_{\sentj}$ inclusive.
    
    \item $\eq$ is a sequence of $0$ or more ``referential equality annotations'', $\eq_1 \ldots \eq_{|\eq|}$. Each member of $\eq$ specifies that some noun phrase within the question refers to the same item in the world as some noun phrase within the sentence $\sent$. 
    
    \item $\ans$ is one or more answer annotations $\ans_1 \ldots \ans_{|\ans|}$. 
    
\end{itemize}

We now describe the form of the $\eq$ and $\ans$ annotations. As a preliminary step, given the paragraph $c$ and sentence $\sent$, we use $\phrases$ to refer to the set of all phrases within $\sent$. Our initial definition of $\phrases$ is 
\[
\phrases = \{(i, j): \senti \leq i \leq j \leq \sentj\}
\]
\commentout{We will refine this further in Section~\ref{sec:bridge}, when we broaden the approach to consider bridging, but for now we will work with this definition.\ml{We have signposting earlier in the section and in section titles, so this sentence is a bit redundant.}}

We also define the set of question phrases $\qphrases$ and passage phrases $\cphrases$ to be
\begin{eqnarray*}
\qphrases &=& \{(i, j): 1 \leq i \leq j \leq m\}\\
\cphrases &=& \{(i, j): 1 \leq i \leq j \leq n\}
\end{eqnarray*}

We can then give the following definitions:

\begin{definition}
Each referential equality annotation $\eq_k$ for $k = 1 \ldots |\eq|$ is a pair $(\phi_k, \pi_k) \in \qphrases \times \phrases$, specifying that the phrase $\phi_k$ in the query refers to the same thing in the world as the phrase $\pi_k$ within $\sent$.
\end{definition}

\begin{definition}
Each answer annotation $\ans_k$ for $k = 1 \ldots |\ans|$ is a pair $(\pi_k, \xi_k) \in \phrases \times \cphrases$  specifying that the answer is given by phrase $\pi_k$, and the full string corresponding to $\pi_k$ after coreference is resolved is the phrase $\xi_k$. If no coreference resolution is required then $\pi_k = \xi_k$.
\end{definition}

To illustrate the treatment of coreference resolution within answers, consider the following: 

$\;$

\noindent
\framebox{\parbox{2.9in}{
\begin{normalsize}
\noindent {\bf Question:} who won wimbledon in 2019

\noindent
{\bf Passage:} Simona Halep is a female tennis player. She won Wimbledon in 2019.
\end{normalsize}}}

$\;$

In this case the single sentence {\em She won Wimbledon in 2019} would be selected by the annotator in step 1, as once coreference is resolved, this entails the answer to the question. The QED annotation would be as follows\:

$\;$

\noindent
{\em who won [=1 wimbledon] in [=2 2019]}

\noindent
{\em [=A She] won [=1 Wimbledon] in [=2 2019]}

$\;$

However, the answer "She" is not sufficient, as it involves an unresolved anaphor. Because of this, the annotator would mark the fact that "She" refers to "Simona Halep" earlier in the passage. In this case the answer is a pair $(\pi, \xi)$ where $\pi$ corresponds to "She" within the sentence, and $\xi$ corresponds to the earlier phrase "Simona Halep". 

\subsection{Extending Annotations to Include Bridging}

\label{sec:bridge}

Bridging anaphora~\cite{clark1975bridging} are frequently encountered in the QA passages in our data, and in Wikipedia more broadly. This section describes an extension to include annotations of bridging anaphora. Consider the following:

$\;$

\noindent
\framebox{\parbox{2.9in}{
\noindent
{\bf Question:} {who won america's got talent season 11}

\noindent
{\bf Passage:} The 11th season of America's Got Talent, an American talent show competition, began broadcasting in the United States during 2016.
Grace VanderWaal was announced as the winner on September 14, 2016.
}}

$\;$

It is clear from context surrounding the sentence "Grace VanderWaal was announced as the winner on September 14, 2016" that the noun phrase "the winner" refers to "the winner of America's Got Talent Season 11", and hence the sentence provides an answer to the question. It is helpful to imagine that there is an implicit prepositional phrase "of America's Got Talent Season 11" modifying "the winner".

Another motivating example is the following:

$\;$

\noindent
\framebox{\parbox{2.9in}{
\begin{normalsize}
\noindent
{\bf Question:} who sang the national anthem at the first game of 2017 world series

\noindent
{\bf Passage:} Game 1 of the 2017 World Series: The ceremonial first pitch was thrown out by members of former Dodger Jackie Robinson’s family, including his widow Rachel. The game marked the 45th anniversary of Robinson’s death. Keith Williams Jr., a gospel singer, performed “The Star-Spangled Banner”, the national anthem.
\end{normalsize}}}

$\;$

In this case it is clear that the sentence "Keith Williams Jr., a gospel singer, performed “The Star-Spangled Banner”, the national anthem" is referring to a performance at Game 1 of the 2017 World Series, and hence that this sentence provides an answer to the question. In some sense there is an implicit prepositional phrase "at the first game of 2017 world series" modifying the entire sentence. 

Recall that the set of phrases within the sentence $s$ was previously defined as
$
\phrases = \{(i, j): \senti \leq i \leq j \leq \sentj\}
$. We extend QED by redefining $\phrases$ to include implicit phrases introduced in the form of implicit prepositional phrases, as in the "winner [of ...]" and "[at the first game ...]" examples above. The modified definition of $\phrases$ includes all phrases of the following form:
(1) Any pair $(i, j)$ such that $\senti \leq i \leq j \leq \sentj$ indicating the subsequence of words $c_i \ldots c_j$ within the sentence.
(2) Any triple $(i, j, p)$ such that $\senti \leq i \leq j \leq \sentj$ and $p$ is a preposition, indicating the implicit noun phrase in the sentence that modifies the phrase $c_i \ldots c_j$ through the preposition $p$.
(3) Any pair $(\hbox{NULL}, p)$ such that $p$ is a preposition, indicating the implicit noun phrase modifying the entire sentence $c_\senti \ldots c_\sentj$ through the preposition $p$.

\section{QED Annotations for the Natural Questions}

We now describe QED annotations over the Natural Questions (NQ) dataset~\cite{47761}. We first describe the annotation process; then describe agreement statistics; finally we describe statistics of  types of referential expression.

\label{sec:annotations}

\begin{figure}[t]
\centering
\input{out-of-qed}
\caption{An example outside of QED's current scope, since multiple passage sentences  contribute an answer.}
\label{fig:out-of-qed}
\end{figure}

We focus on questions in the NQ corpus that have both a passage and short answer marked by the NQ annotator. We exclude examples where the passage is a table. A QED annotator was presented with a question/paragraph pair. In a first step they determine whether: (1) there is a valid short answer within the paragraph (note that they can overrule the original NQ judgment), and there is a valid QED explanation for that answer; (2) there is a valid short answer within the paragraph, but there is no valid QED explanation for that answer. (See Figure~\ref{fig:out-of-qed} for a representative example in this category, in which multiple sentences are required to justify an answer, thus violating the single-sentence assumption of QED); (3) there is no valid short answer within the passage (hence the original NQ annotation is judged to be an error). 10\% of all examples fell into category (3). Of the remaining 90\% of examples which contained a correct short answer, 77\% fell into category (1), and 23\% fell into category (2).

Three QED annotators\footnote{Three of the authors of this paper.}
annotated 7638 training examples (5154/1702/782 in categories 1/2/3 respectively), and 1353 dev examples (1019/183/151 in categories 1/2/3).

\subsection{Agreement Statistics}

Each of the three annotators marked a common set of 100 examples drawn from the development set. Average accuracy of classification of instances was 73.9.\%\footnote{One annotator was more conservative interpreting the single sentence assumption. Pairwise accuracy breakdown was thus 81.2/72.3/68.1\%. Given the high number of ``debatable" instances reported in the Natural Questions paper, this divergence is however unsurprising.} Average pairwise F1 on mention identification/mention alignment, conditioned on both annotators labeling instances as amenable to QED, was 88.4 and 84.1 respectively.

\begin{table}[t]
\small
    \centering
    \begin{tabular}{l|cccc}
\toprule
 & \multicolumn{4}{c}{Referential Link Count} \\
  & 0 & 1 & 2 & 3 \\
 \midrule
 Instances & 54 & 649 & 294 & 6 \\
 \bottomrule
    \end{tabular}
    \caption{Referential link count frequency distribution in a random sample of 1000 instances.}

    \label{tab:link_distribution}
\end{table}

\subsection{Types of Referential Expressions}

\begin{figure*}[t!]
\centering

\begin{tabular}{p{.97\textwidth}} \toprule
  \footnotesize
                        \footnotesize\textbf{\underline{Pronominal reference}}\\
                        \footnotesize \textbf{Question}: how many blocks in \spanone{the great pyramid of giza}{1} \\
                        \footnotesize \textbf{Wikipedia page}: Great\_Pyramid\_of\_Giza\\ 
                        \footnotesize \textbf{Passage}: Based on these estimates, building the pyramid in 20 years would involve installing approximately 800 tonnes of stone every day. \spansent{Additionally, since }\spanone{\textbf{it}}{1}\spansent{consists of an estimated }\spanans{2.3 million}{\textsc{a}}\spansent{blocks, completing the building in 20 years would involve moving an average of more than 12 of the blocks into place each hour, day and night.} The first precision measurements of the pyramid were made by Egyptologist Sir Flinders Petrie in 1880--82 and published as The Pyramids and Temples of Gizeh. Almost all reports are based on his measurements[...] \commentout{Many of the casing stones and inner chamber blocks of the Great Pyramid fit together with extremely high precision. Based on measurements taken on the northeastern casing stones , the mean opening of the joints is only 0.5 millimetre wide (1/50 of an inch).}
\\ \midrule
\footnotesize
                        \footnotesize\textbf{\underline{Inexact match}} \\ 
                        \footnotesize \textbf{Question}: where does \spanone{\textbf{the term sixes and sevens}}{1} originate \\
                        \footnotesize \textbf{Wikipedia page:} At\_sixes\_and\_sevens \\
                        \footnotesize \textbf{Passage}: \spanans{An ancient dispute between the Merchant Taylors and Skinners livery companies}{\textsc{a}}\spansent{ is the probable origin of}\spanone{\textbf{the phrase}}{1}\spansent{.} The two trade associations, both founded in the same year (1327), argued over sixth place in the order of precedence. In 1484, after more than a century and a half of bickering, the Lord Mayor of London Sir Robert Billesden ruled that at the feast of Corpus Christi, the companies would swap between sixth and seventh place and feast in each other's halls[...]\commentout{Nowadays, they alternate in precedence on an annual basis.}
\\ \midrule
                        \footnotesize\textbf{\underline{Answer bridging/coref}} \\
                        \footnotesize \textbf{Question}: what is \spanone{whitney houston's mother}{1}'s name \\
                        \footnotesize \textbf{Wikipedia page}: Cissy\_Houston \\
                        \footnotesize \textbf{Passage}: \footnotesize \spanans{\textbf{Emily ``Cissy'' Houston}}{\textsc{a}} (née Drinkard; born September 30, 1933) is an American soul and gospel singer. After a successful career singing backup for such artists as Dionne Warwick, Elvis Presley and Aretha Franklin, Houston embarked on a solo career, winning two Grammy Awards for her work.\spansent{\textbf{Houston} is}\spanone{the mother of singer Whitney Houston}{1}\spansent{, grandmother of Whitney's daughter, Bobbi Kristina Brown, aunt of singers Dionne and Dee Dee Warwick, and a cousin of opera singer Leontyne Price.}
\\ \midrule
                        \footnotesize\textbf{\underline{Entity bridging}}\\
                        \footnotesize \textbf{Question}: who sang \spanone{the national anthem}{1} at \spantwo{\textbf{the first game of 2017 world series}}{2} \\
                        \footnotesize \textbf{Wikipedia page}: 2017\_World\_Series \\
                        \footnotesize \textbf{Passage}:
\textbf{Game 1}: The ceremonial first pitch was thrown out by members of former Dodger Jackie Robinson's family, including his widow Rachel. The game marked the 45th anniversary of Robinson's death, and the 2017 season was the 70th anniversary of his breaking of the baseball color line. \spantwo{\textbf{[$\dots$]}}{2}\spanans{Keith Williams Jr.}{\textsc{a}}\spansent{, a gospel singer, performed }\spanone{``The Star-Spangled Banner", the national anthem}{1} \spansent{.}
\\ \midrule
                        \footnotesize\textbf{\underline{Generic reference}} \\
                        \footnotesize \textbf{Question}: what is the function of \spanone{\textbf{a paints binder}}{1}\\
                        \footnotesize \textbf{Wikipedia page}: Paint \\
                        \footnotesize \textbf{Passage}: \spanone{\textbf{The binder}}{1}\spansent{is the}\spanans{film-forming}{\textsc{a}}\spansent{component of paint.} It is the only component that is always present among all the various types of formulations. Many binders are too thick to be applied and must be thinned. The type of thinner, if present, varies with the binder.
\\
\bottomrule
\end{tabular}

\caption{\label{tab:annotatedExamples}
Examples from the QED dataset, grouped according to different types of referential equalities.}
\label{fig:bigexamples}
\end{figure*}

\begin{figure}[h]
    \centering
    \begin{footnotesize}
    \begin{tabular}{|p{1.1in}|p{1.5in}|}
    \hline
    Question Expression&Passage Expression\\
    \hline
    \hline
    how i.met your mother&the CBS television sitcom How I Met Your Mother\\
    \hline
    the most wins in the nfl&most wins\\
    \hline
    mantis&Mantis\\
    \hline
    the nashville sound&
    Countrypolitan - a smoother sound typified through the use of lush string arrangements with a real orchestra and often, background vocals provided by a choir\\
    \hline
    a permit driver&a driver operating with a learner 's permit\\
    \hline
    god's not dead a light in the darkness&it\\
    \hline
    the current president of un general assembly&
    the United Nations General Assembly President of its 72nd session beginning in September 2017\\
    \hline
    the new maze runner movie&
    Runner : The Death Cure\\
    \hline
    a box lacrosse team&
    a team\\
    \hline
    \end{tabular}
    \end{footnotesize}\vspace{-4pt}
    \caption{Referential equalities from the QED corpus.}
    \label{fig:refexamples}
\end{figure}

\begin{figure}[t]
    \centering
    \begin{footnotesize}
    \resizebox{\columnwidth}{!}{%
\begin{tabular}{|l||c|c|c|c|c|c|c||c|}
\hline
\backslashbox{Qu.}{Ps.}&P&N&A&G&Pn&B&M&T\\   
\hline
\hline
Proper&44&0&16&0&9&4&0&73\\
\hline
Def.&4&6&4&0&0&1&1&16\\
(Non-Ana)&&&&&&&&\\
\hline
Def.&0&1&1&0&0&0&0&2\\
(Ana)&&&&&&&&\\
\hline
Generic&0&0&0&6&0&0&0&6\\
\hline
Pronoun&0&0&0&0&0&0&0&0\\
\hline
Bridge&0&0&0&0&0&0&0&0\\
\hline
Misc&2&0&0&0&0&0&1&3\\
\hline
\hline
Total&50&7&21&6&9&5&2&100\\
\hline
\end{tabular}
    }%
\end{footnotesize}
    \caption{Counts for 100 randomly drawn referential equality annotations from the QED corpus, subcategorized by expression type in the question (Qu.) and passage (Ps.). P/N/A/G/Pn/B/M refer to Proper/Def(non-ana)/Def(ana)/Generic/Pronoun/Bridge/Misc.}
    \label{fig:rtypes}
\end{figure}

The referential equality annotations are a major component of QED. Figure~\ref{fig:bigexamples} shows some full QED examples from the corpus, and  Figure~\ref{fig:refexamples} shows some example equalities from the corpus. In this section, in an effort to gain insight about the types of phenomena present, we describe statistics on types of referential equalities. We subcategorize referring expressions into the following types:\footnote{For formal discussion, see \cite{carlson1977unified,krifka2003bare,abbott2004definiteness, mikkelsen2011copula} among others.}

\paragraph{Proper Names} Examples are ``How I met your Mother'' or ``the cbs television sitcom how i met your mother''.

\paragraph{Non-Anaphoric Definite NPs} These are expressions such as ``the president of the United States'' or ``the next Maze Runner film''. The majority involve one or more common nouns (e.g., "president", "film") together with a proper name, thereby defining a new entity that is in some sense a "derivative" of the underlying proper name.

\paragraph{Anaphoric Definite NPs} These are definite NPs, most often from within the passage rather than the question, that require context to be interpreted. Examples are "the series" referring to an earlier mention of "the Vampire Diaries" within the passage, or  "the winner" referring to "the winner of America's got Talent Season 11".

\paragraph{Generics} Examples are "a dead zone" in the question "what causes a dead zone in the ocean", or "Dead zones" in the passage sentence "Dead zones are low-oxygen areas caused by ...".

\paragraph{Pronouns} Examples are it, they, he, she.

\paragraph{Bridging} Referential expressions in the passage sentence that use bridging  (see Section~\ref{sec:bridge}). 

\paragraph{Miscellaneous} All referential expressions not included in the categories above.

Table~\ref{tab:link_distribution} shows the frequency distribution of per-instance referential equality counts. Figure~\ref{fig:rtypes} shows an analysis of 100 referential equality annotations from QED, with a breakdown by type of  referring expression in the question and passage. Proper names, non-anaphoric definites, and generics dominate expression types in the question (73, 16, and 6 examples respectively). Expressions in the sentence are more diverse, with a much greater proportion of anaphoric definites, pronouns, and bridging examples (21, 9, and 5 cases respectively).

Finally, as an indication of the difficulty of the referential equality task, we note that in only 12\% of all referential equalities in the 100 examples in Figure~\ref{fig:rtypes} is there an exact string match (after lower-casing of both question and passage) between the question and passage referential expression. 

\commentout{
44 proper proper
  16 proper a-def
   9 proper pronoun
   6 generic generic
   5 nona-def nona-def
   4 proper bridge-sent
   4 nona-def proper
   4 nona-def a-def
   2 misc proper
   1 nona-def nona-def    
   1 nona-def misc
   1 nona-def bridge-sent
   1 misc misc
   1 a-def nona-def
   1 a-def a-def
}

\commentout{
\paragraph{Indefinite NPs} These are indefinite NPs that are not generics.
\paragraph{Dates or Years} For example "March 30, 2018", or "2018".
}

\section{Tasks and Baseline Results}
\label{sec:tasks_and_baselines}
We release the QED dataset with the intention to spur research into QED-based tasks and models. In this section, we introduce four potential modeling tasks using the data and describe baseline approaches and results for the first two tasks.

\subsection{Four Tasks}

Each QED example is a $(q, d, c, a, e)$ tuple where $q$ is a question from the NQ corpus, $d$ is a Wikipedia page, $c$ is a long answer (typically a paragraph) within $d$, $a$ is a short answer within $c$, and $e$ is a QED explanation. We use ${\cal E}$ to refer to set of evaluation examples (either the development or test set). 

Such data could potentially be used in many different ways. We highlight the following four tasks, in  order of increasing complexity:

    \paragraph{Task 1} Given a $(q, d, c, a)$ 4-tuple, make a prediction $\hat{e} = f(q, d, c, a)$ where $f$ is a function that maps a $(q, d, c, a)$ tuple to an explanation. We might for example define $f(q, d, c, a) = \arg\max_e p(e | q, d, c, a; \theta)$ under some model $p(\ldots)$. The evaluation measure is then
    \[
\frac{1}{|{\cal E}|}    \sum_{(q, d, c, a, e) \in {\cal E}} l_1(e, f(q, d, c, a))
    \]
    where $l_1(e, \hat{e})$ is a per-example evaluation measure indicating how close $\hat{e}$ is to $e$.
    
        \paragraph{Task 2} Given a $(q, d, c)$ triple, predict $(\hat{a}, \hat{e}) = f(q, d, c)$, where $f$ is a function that maps a $(q, d, c)$ pair to a short-answer/explanation triple.  We might for example define $f(q, d, c) = \arg\max_{a, e} p(a, e | q, d, c; \theta)$ under some model $p(\ldots)$. The evaluation measure is
    $
    \sum_{(q, d, c, a, e) \in {\cal E}} l_2((a, e), f(q, d, c))
    $
    where $l_2$ is some per-example measure. 
    
    \paragraph{Task 3} Given a $(q, d)$ pair, predict $(\hat{c}, \hat{a}, \hat{e}) = f(q, d)$, where $f$ is a function that maps a $(q, d)$ pair to a long-answer/short-answer/explanation triple.  We might for example define $f(q, d) = \arg\max_{c, a, e} p(c, a, e | q, d; \theta)$ under some model $p(\ldots)$. The evaluation measure is
    $
    \sum_{(q, d, c, a, e) \in {\cal E}} l_3((c, a, e), f(q, d))
    $
    where $l_3$ is some per-example measure. 
    \paragraph{Task 4} As in Task 3, given a $(q, d)$ pair, predict $(\hat{c}, \hat{a}, \hat{e}) = f(q, d)$. One part of the evaluation is the same as in Task 3. But in addition, we require the explanations generated by $f(\ldots)$ to be {\em faithful} with respect to the reasoning process of the underlying model. This will require an evaluation measure for faithfulness, which is an open question beyond the scope of this paper.

Accurate models for Tasks 1, 2, and 3 even if they do not generate faithful explanations (Task 4), may still have considerable utility. However, faithful models have several desirable characteristics (see Section~\ref{sec:discuss}); we view them as a major avenue for future work.

In the remainder of this section we describe results for baseline models on Tasks 1 and 2. The intention here is to establish baseline results as a reference point for future work on QED models and to get an idea of tractability of recovery of QED annotations.

\label{sec:baseline}

\subsection{A Baseline Model for Task 1}

Our baseline model for Task 1 is an extension of the recently proposed coreference resolution model of~\citet{Joshi2019SpanBERTIP} and~\citet{Lee2017EndtoendNC}. We present two variations on the model, the first trained on coreference data alone, the second trained on coreference data with fine-tuning on QED annotations

\subsubsection{The Coreference Resolution Model}

We give a brief recap of the approach of ~\citet{Joshi2019SpanBERTIP} and~\citet{Lee2017EndtoendNC}. Given some document $d$ and a candidate mention $x$, corresponding to a span within $d$, define ${\cal Y}(x)$ to be the set of potential antecedents for $x$. Each antecedent is either a span in the document with start-point before $x$ in the document, or $\epsilon$ signifying that $x$ does not have an antecedent. We can then define a distribution over the antecedent spans ${\cal Y}(x)$ as 
    $p(y| x, D) = \frac{e^{s(x, y)}}{\sum_{y' \in {\cal Y}(x)} e^{s(x, y')}}$
where
\begin{align*}
    s(x, y) &= 
    \left\{ \begin{array}{ll}
         0 & \mbox{\hspace{-0.5cm} if $y = \epsilon$};\\
         s_m(x) + s_m(y) + s_c(x, y)& \mbox{o.t.} \end{array} \right. \\
    s_m(x) &= \texttt{FFNN}_m({g_{x}}) \\
    s_c(x, y) &= \texttt{FFNN}_c({g_{x}}, {g_{y}})
\end{align*}
where $g_{x}$ and $g_{y}$ are span representations obtained by concatenating the SpanBERT representation of the first and last token in each mention span. The scoring functions $s_m$ and $s_c$ represent mention and joint span match scores respectively.

\citet{Lee2017EndtoendNC} describe a method for training the model based on log-likelihood, and a beam search method that uses the scores $s_m(\ldots)$ to filter mentions and antecedents. The final output from the model is a hard clustering of the potential mentions into coreference clusters.

\subsubsection{The Model Applied to Task 1}

Assume an example contains a question $q$ of $\nq$ tokens $q_1 \ldots q_{\nq}$ and a passage $c$ consisting of $\nc$ tokens $c_1 \ldots c_{\nc}$. We denote the title of the Wikipedia page separately as the sequence $t$ of $k$ tokens $t_1 \ldots t_{k}$. The model considers the concatenation of these token sequences, \[{\small \texttt{[CLS]} t_1 \ldots t_{k}  \texttt{[S1]}  q_1 \ldots q_{\nq} \texttt{[S2]}   c_1 \ldots c_{\nc} \texttt{[SEP]},}\] as an input document.\footnote{We simply use [S1] = "." and [S2] = "?" as separators.} The model is tasked with predicting the referential equality annotations $e = e_1 \ldots e_k$ in the QED annotation. \commentout{For now we do not attempt to predict the answer annotation $a$.} We do assume that the NQ short answer is also an input to the model, used to restrict the position of referential equality annotations in the passage; we describe this restriction  below.

QED referential equality annotations are of two types: (1) coreferential links between noun phrases in the question and in the passage, and (2) coreferential links between a noun phrase in the question and an implicit argument in the passage. We observe that many implicit arguments link to the title of the passage, so we model the latter annotation type as a coreferential link between the question mention and the title span $t_1 \ldots t_{k}$. In the untrained baseline, we restrict $s_m$ to only score mentions in the sentence containing the answer. In both models we restrict $s_c$ to only score coreferential links between the query and the passage or between the query and the title (all other values for $s_m$ or $s_c$ are set to $- \infty$).

We finally post-process the cluster outputs as follows: for each cluster we output the first cluster mention in the question paired with the first cluster mention in the passages. If there is no cluster mention in the passage, then we output the question mention paired with an implicit argument.

For the untrained baseline, we did not use expert annotated QED data but instead used the CoNLL OntoNotes coreference dataset~\cite{Pradhan2012CoNLL2012ST} to train the pretrained SpanBERT model. For the fine-tuned baseline, we further trained the model with the training portion of QED data converted into coreference format. We used SpanBERT ``large'', with a maximum span width of 16 tokens, a top span ratio of 0.2, 30 max antecedents per mention. In fine-tuning, we used an initial learning rate of $3 \cdot 10^{-4}$ and trained for 3 epochs on the QED training set.


    
    
    

\begin{table}
\footnotesize
    \centering
    \resizebox{\columnwidth}{!}{%
    \begin{tabular}{@{} l|ccc|ccc @{}}
    \toprule
     &\multicolumn{3}{c|}{Mention } &  \multicolumn{3}{c}{Mention }\\
     &\multicolumn{3}{c|}{ Identification} &  \multicolumn{3}{c}{ Alignment}\\
     & P & R & F1 & P & R & F1 \\
    \midrule
    \hspace{2pt} zero-shot & 59.0 & 35.6 & 44.4 & 47.7 & 28.8 & 35.9 \hspace{2pt} \\  
    \hspace{2pt} fine-tuned & 76.8 & 68.8 & 72.6 & 68.4 & 61.3 & 64.6 \hspace{2pt} \\  
     \bottomrule
    \end{tabular}
    }%
    \caption{SpanBERT model performance for Task 1: recovering QED annotations when the correct answer is given.}
    \label{tab:baseline_performance}
\end{table}

We evaluate both mention identification (the identification of individual referential expressions in the question and passage) and referential equality detection (the identification of pairs of referential expressions). We compute precision, recall, and F1 measure in both cases. Evaluation results are reported in Table~\ref{tab:baseline_performance}. The table shows results for both the zero-shot model, trained on coreference data alone, and a fine-tuned model, which is fine-tuned on QED annotations.\footnote{Official evaluation code will be released with the dataset.}

\subsection{A Baseline Model for Task 2}

Our baseline model for Task 2 is a straightforward extension of the baseline model for Task 1. We build a model of the form
\begin{eqnarray*}
&&p(a, e| q, d, c; \theta)\\
&=& p^{(1)}(a | q, d, c; \theta^{(1)}) p^{(2)}(e | a, q, d, c; \theta^{(2)})
\label{eq:condmodel}
\end{eqnarray*}
where $p^{(1)}$ is an existing QA model (similar to \newcite{alberti2019bert}), and $p^{(2)}$ is the baseline model for Task 1. Thus we simply compose an existing question-answering model with an answer agnostic model that recovers explanations.

The answer scoring component of the model computes answer candidate representations $g_z$ in the same way as the Task 1 baseline computes mention representations. The score of an answer $z$ is then computed as
\begin{align*}
    s_a(z) &= \texttt{FFNN}_a({g_z}).
\end{align*}
Mention representations are shared between $p^{(1)}$ and $p^{(2)}$, so the only new parameters belong to a single hidden layer feed-forward net $\texttt{FFNN}_a$ that computes the answer score for each mention. No further dependence is introduced between the answer and explanation predictions. We train $p^{(1)}$ and $p^{(2)}$ in a multitask fashion, by minimizig the weighted sum of the question answering and coreference cross entropy losses. Our best results are obtained with a weight of 5 on the coreference loss and 2 epochs of training. The best answer accuracy and QED F1 are obtained for different base learning rates of $2 \cdot 10^{-5}$ and $5 \cdot 10^{-5}$ respectively.

\subsubsection{Results}

\begin{table}
\footnotesize
    \centering
    \resizebox{\columnwidth}{!}{%
    \begin{tabular}{@{} l|ccc|ccc|c @{}}
    \toprule
     &\multicolumn{3}{c|}{Mention } &  \multicolumn{3}{c|}{Mention } & Answer \\
     &\multicolumn{3}{c|}{ Identification} &  \multicolumn{3}{c|}{ Alignment} & Accuracy\\
     & P & R & F1 & P & R & F1 & \\
    \midrule
    
    \hspace{2pt} QED-only & 74.1 & 63.8 & 68.6 & 63.6 & 54.9 & 58.9 & - \\
    \hspace{2pt} QA-only & - & - & - & - & - & - & 73.4 \\
    \hspace{2pt} QA+QED & 77.5 & 64.6 & 70.5 & 68.6 & 57.3 & 62.4 & 74.5 \\
    
     \bottomrule
    \end{tabular}
    }%
    \caption{SpanBERT model performance for Task 2: recovering answer and QED annotations given a passage that is known to contain the answer.}
    \label{tab:baseline_performance_no_answer}
\end{table}

In Table~\ref{tab:baseline_performance_no_answer} we report results for Task 2 for three separate variations of the approach described in the previous section. QED-only fine-tunes  $p^{(2)}$ on the QED training set only. QA-only fine-tunes $p^{(1)}$ on all the paragraphs of the NQ dataset that contain a short answer. QA+QED fine-tunes both $p^{(1)}$ and $p^{(2)}$ on all NQ and QED data. We obtain the encouraging result that both QA and QED metrics improve significantly in the final multitask setting, despite the fact that the QED training data (5154 examples) amounts to only 6\% of the data available for QA (91632 examples).
\section{Rater Study}
\label{sec:study}

A system which makes use of QED explanations to answer a question is one which decomposes its reasoning process into human-interpretable chunks. We hypothesize that exposing QED explanations should improve a user's ability to spot errors made by an automated QA system. To this end, we evaluate QED explanations using a rater study. 
\ec{we should add a screenshot of UI, either here or in appendix}\ml{Agreed, but we're tight on space. Maybe this can be addressed when they give us extra pages?}
\subsection{Task Setup}
\label{sec:study-method}
Given a question, passage, and a candidate answer span,  raters were tasked with assessing whether the candidate answer was correct or incorrect, and indicating the confidence of their assessment.

We obtained the data for the study by taking a random set of 50 correct answers and 50 incorrect guesses from the NQ baseline model on the Natural Questions dev set. So as to ensure that the task was sufficiently challenging, correct instances were the \emph{gold} answer spans on question/passage pairs where the model produced a false negative.\footnote{That is, where an answer existed in the passage, but the model was not confident about it.} Incorrect instances were false positive guesses from the model.

A total of 354 raters, all of whom are US-residents and native English speakers, were divided into three disjoint pools to perform the task in three distinct test settings:
The \textbf{None} group of raters (n=121) was presented with a question, passage, and a highlighted answer span.
The \textbf{Sentence} group (n=117) was provided with additional highlighting of the sentence containing the answer, with no distinction made between referential equalities and predicates.
The \textbf{QED} group (n=116) was provided with additional highlighting to indicate referential equalities between spans in the question and spans in the passage. 
On average, a given rater provided judgments for 41 questions.

In each case, raters were told that highlighting was the output of ``an automated question answering system'' that was incorrect ``about half of the time.'' Where explanations were present, they were manually imputed to simulate the inferences of a hypothetical model that used a QED-style reasoning process. Additionally, raters were told that the system made use of the highlighted information to produce its candidate answers. 

\subsection{Results}

\begin{table}[t]
\small
    \centering
    \begin{tabular}{@{} l|ccc|c @{}}
    \toprule
    &\multicolumn{3}{c|}{Accuracy} & F1 \\
     & All & Corr & Incorr/Pred/Ref & Incorr\\
    \midrule
    None & 67.5 & 90.4 & 44.3/43.9/44.7 & 57.6 \\ 
    Sentence & 69.7 & \textbf{92.4} & 47.1/46.1/48.0 &  60.9 \\
    QED & \textbf{70.2} & 90.6 & \textbf{49.7}/\textbf{48.2}/\textbf{51.0} &  \textbf{62.5} \\
    \bottomrule
    \end{tabular}
    \caption{Rater study results. Corr\ and Incorr\ are accuracies of raters in each group on correct and incorrect instances respectively, with incorrect instances further broken into Pred(icate) and Ref(erence) model errors. F1 is on the task of identifying incorrect instances.}
    \label{tab:rater_acc}
\end{table}

\begin{figure*}[t]
    \centering
    \includegraphics[width=6in]{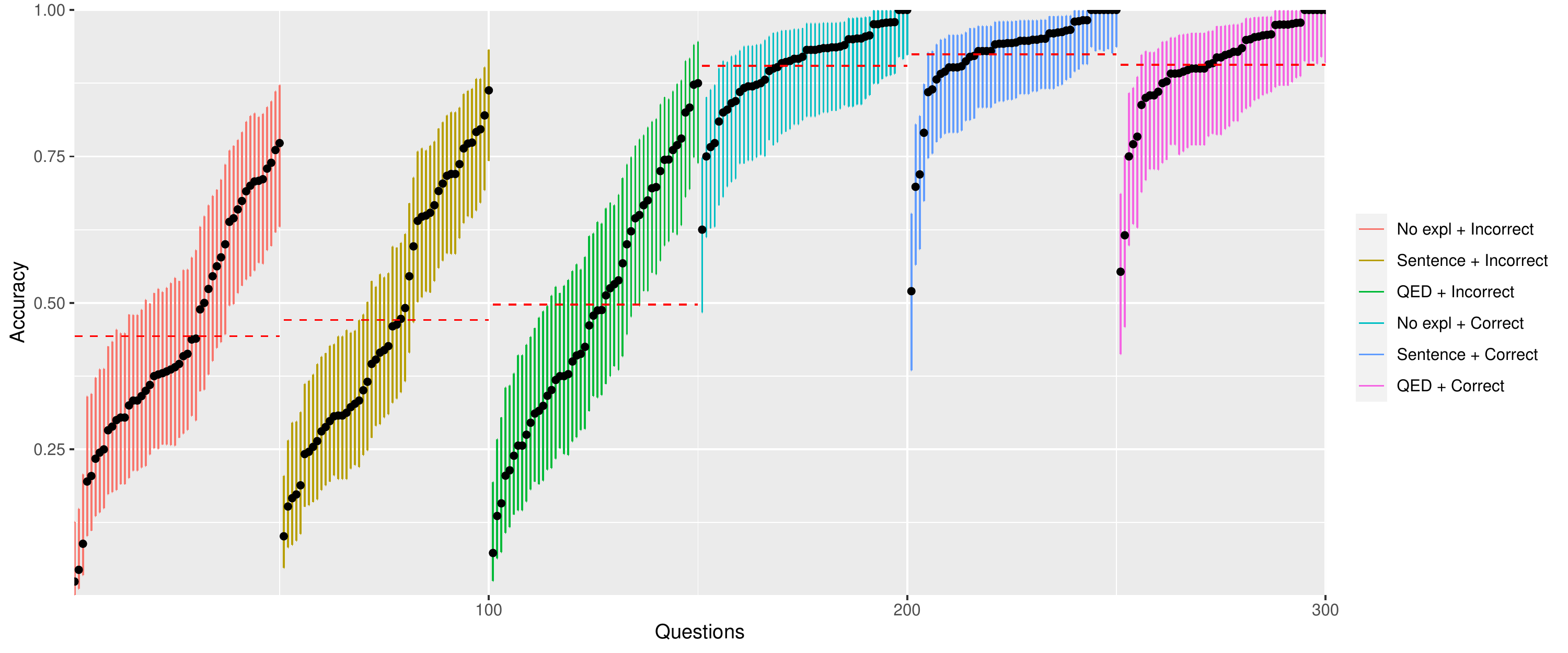}
    \caption{Sorted, per-question evaluation accuracies from different rater study settings, with 95\% binomial confidence intervals.
    Left three plots correspond to trials with incorrect answers highlighted; right three plots to trials with correct answers highlighted.
    Dashed red lines correspond to the average accuracy for each setting, identical to the numbers in Table~\ref{tab:rater_acc}.}
    \label{fig:question_dispersion}
\end{figure*}

Average rater accuracies for each test setting are presented in Table~\ref{tab:rater_acc}. We see that, in aggregate, QED explanations improved accuracy on the task over and above the other test settings, and gave the most improvement on the identification of answers that were incorrect. These improvements translate to incorrect answers resulting from both predicate and reference model errors.

Somewhat surprisingly, highlighting just the sentence containing the answer improved accuracy more than including referential equality highlighting on instances that were correct. This is likely because raters' propensity to mark instances correct decreases as the complexity of explanations increases, from None (73.1\%) to Sentence (72.6\%) to QED (70.5\%).

Also clear from Table~\ref{tab:rater_acc} is that rater accuracy is much lower on incorrect instances. Even though raters were told that the answers presented were incorrect half of the time, they marked the model guess as correct roughly 71\% of the time.\footnote{While this confirmation bias presents an interesting challenge for future work, it is not a shortcoming of our results: Raters were not trained to do well on the task, as we aimed to approximate how users interact with automated QA systems.}

Figure~\ref{fig:question_dispersion} provides another perspective on the disparity in judgments on correct/incorrect instances summarized in Table~\ref{tab:rater_acc}. The instances receiving the highest accuracy in the incorrect pool are harder for raters on average than most of the correct instances, and the lowest accuracy on incorrect instances is far lower than that of any of the correct instances. 
The wide distribution of accuracies on incorrect instances ($\sigma$$\approx$0.50) seen in Figure~\ref{fig:question_dispersion} was also reflected in the rater pool ($\sigma$$\approx$0.45).
The challenging nature of incorrect instances speaks to the promise of improvements from QED explanations.

\commentout{In addition to being harder for the raters, the \textit{detection of incorrect instances} is a useful way to re-frame the task.
We computed F1 scores for the test settings and found that adding annotations helped. The None, Sentence, and QED settings had F1 scores of 57.6\%, 60.9\%, and 62.5\%, respectively.\ec{potentially cut this paragraph}\da{Pr maybe add to the table? Emily thought the F1 angle was interesting.}\ml{Added to table.}}

\subsection{Effectiveness of explanations}

How statistically significant are the results reported in Table~\ref{tab:rater_acc}?
The 14,115 test instances were spread across 354 raters and 100 questions.
To control for the correlations induced by the rater and question groups, we fit a generalized linear mixed model (GLMM) using the \texttt{rstanarm} R package~\cite{rstanarm}.
We used the formula
$
\texttt a \, \mathtt \sim \, \texttt{c * e + (1|r) + (1|q)},
$
where \texttt{a} is whether or not the rater accurately marked the instance; \texttt{c} is whether the instance was Correct or Incorrect; \texttt{e} is the explanation test setting of None, Sentence, or QED; \texttt{r} is the rater id; and \texttt{q} is the question id.
This formula specifies a regression of the log-odds of the rater accuracy on the fixed effects of instance correctness (\texttt{c}) and explanation setting (\texttt{e}), while allowing for random effects in the raters (\texttt{r}) and questions (\texttt{q}).
Ultimately we are interested in the magnitude and statistical properties of \texttt{e} under the various test settings.
\begin{table}[t]
    \small
    \centering
    \begin{tabular}{lc}
    \toprule
    Parameter & Coefficient (SD) \\
    \midrule
    (Intercept) & -0.31 (0.15) \\
    + Incorrect+Sentence & 0.15 (0.11) \\
    + Incorrect+QED & 0.25 (0.11) \\
    + Correct+None & 2.94 (0.21) \\
    + Correct+Sentence & 3.04 (0.13) \\
    + Correct+QED & 2.69 (0.13) \\
    \bottomrule
    \end{tabular}
    \caption{Generalized linear mixed model fixed effect coefficients, showing mean and standard deviation of 10k MCMC samples.
    The Intercept corresponds to the Incorrect+None setting.}
    \label{tab:glmm_coeffs}
\end{table}

Table~\ref{tab:glmm_coeffs} shows the fixed effect coefficient and standard deviations for each setting. 
The presence of QED explanations in the Incorrect setting increased the log-odds of rater accuracy by 0.25, with a posterior predictive p-value of 0.015 that this effect is greater than zero.
The comparable effect for Sentence explanations was 0.15, with a posterior predictive p-value of 0.08.\ec{is there a reason not putting these p values in table 4?} The rater and question random effects had standard deviations of 0.63 and 0.90 respectively, reflecting again the high variance of questions shown in Figure~\ref{fig:question_dispersion}.

As we saw earlier, the effects of explanations in the Correct setting was reversed:
the Sentence explanations caused a small, statistically insignificant increase in log-odds, while QED explanations caused a statistically significant drop in log-odds.

\section{Discussion}
\label{sec:discuss}
\subsection{QED and strong explainability}

It is an open question as to what constitutes a good explanation~\cite{lipton2001good}. A major inflection point in the discussion is the notion of \textit{faithfulness}~\cite{ross2017right, lipton2016mythos}. We say a model's explanations are faithful when there is a causal relationship between an explanation and a prediction. That is, when  an explanation changes, the outputs change accordingly. When this is not true, we say a model generates \textit{rationales}, which have the appearance of justifying its outputs, but without causal guarantees~\cite{ehsan2018rationalization}.

While the models described in Section~\ref{sec:tasks_and_baselines} fall into the latter category, we believe QED is a promising framework for strongly explainable QA. This is due in large part to its commitment to the cognitive reality of reference and entailment. We can say, definitively, that in order for a sentence to answer a question about a thing, its meaning must involve that thing in a very particular sense. Posed counterfactually, when you break referential equality, you break answerhood, and the same argument follows for predicate entailment. Unlike other intelligent behavior that may permit of post-hoc rationalization at best \cite{ehsan2019automated}, certain forms of high-level linguistic reasoning are in fact amenable to strong explanation.

\subsection{Potential Extensions to the QED Framework}

QED exists in between relatively unstructured explanation forms on the one hand, such as attention distributions~\cite{wiegreffe2019attention,jain2019attention,mohankumar2020transparent} or sequential outputs~\cite{camburu2018snli, camburu2019make, narang2020wt5, kumar2020nile} and more elaborate, discrete semantic representations that can in theory be applied to explainable QA~\cite{abzianidze2017parallel, Wolfson_2020}.

The version of QED presented here is a broad coverage, yet limited instantiation of a framework, in which explanations are semantic relations whose substructures are defined in terms of formally motivated linguistic categories. However, in keeping with its modularity, we can extend QED to account for these by looking to semantic relations beyond referential equality and predicate entailment, such as set-membership noun phrase~\cite{hearst1992automatic} and interclausal~\cite{miltsakaki2004penn, lamm2018textual, tandon2019wiqa} relations.

\subsection{Future uses of QED representations}

\label{sec:future}

Our hope is that QED representations may be useful in a variety of extensions to extant QA systems. Some examples are as follows:

\paragraph{Ambiguous Questions.} Consider again the question in Figure~\ref{fig:intro-example}, "who wrote the film howl’s moving castle". Now consider the question "who wrote howl’s moving castle". In this case there are two possible answers, depending on whether the author of the question is referring to the film or novel. It would be natural for a system to provide two possible answers~\cite[see, e.g.][]{min2020ambigqa}, with two possible QED explanations highlighting the differing assumptions underlying each answer. Such referential ambiguities are common, and the centrality of referential equality in QED annotations should mean that they are useful in this scenario.

\paragraph{Complex Referential Equalities.} Consider the question "meaning of whiskey in the jar by metallica". The Wikipedia page for "Whiskey in the Jar" says the following:

$\;$

\noindent
\framebox{\parbox{2.9in}{
\begin{normalsize}
{\bf Passage:}
    "Whiskey in the Jar" is an Irish traditional song set in the southern mountains of Ireland. The song, about a rapparee (highwayman) who is betrayed by his wife or lover, is one of the most widely performed traditional Irish songs and has been recorded by numerous artists since the 1950s.
\end{normalsize}}}

$\;$

\noindent
A good answer could be that the song is "about a rapparee $\ldots$ who is betrayed by his wife or lover", assuming that the Metallica song is a variant of the Irish traditional song. Thus the validity of this answer hinges on a complex referential equality, between the Metallica version and the original. Examples that require this type of complex referential reasoning are quite common, and the centrality of reference in QED should be relevant.

\section{Conclusions}

We have described QED, a framework for explanations in question answering, and we have introduced a corpus of QED annotations. The framework is grounded in referential equality, and entailment. In addition we have described baseline models for two QED-based tasks, and a rater study utilizing QED annotations.

Future work should consider the development of  models that provide faithful explanations based on QED; extensions of QED, for example to handle multi-sentence inference or referential phenomena going beyond equality; and applications of QED, for example to sentences with multiple potential answers, to questions that are vague or underspecified, or to referential equalities that require significant inference to be justified.

\bibliography{references}

\begin{thebibliography}{39}
\expandafter\ifx\csname natexlab\endcsname\relax\def\natexlab#1{#1}\fi

\bibitem[{Abbott(2004)}]{abbott2004definiteness}
Barbara Abbott. 2004.
\newblock Definiteness and indefiniteness.
\newblock \emph{The handbook of pragmatics}, 122.

\bibitem[{Abzianidze et~al.(2017)Abzianidze, Bjerva, Evang, Haagsma, Van~Noord,
  Ludmann, Nguyen, and Bos}]{abzianidze2017parallel}
Lasha Abzianidze, Johannes Bjerva, Kilian Evang, Hessel Haagsma, Rik Van~Noord,
  Pierre Ludmann, Duc-Duy Nguyen, and Johan Bos. 2017.
\newblock The parallel meaning bank: Towards a multilingual corpus of
  translations annotated with compositional meaning representations.
\newblock \emph{arXiv preprint arXiv:1702.03964}.

\bibitem[{Alberti et~al.(2019)Alberti, Lee, and Collins}]{alberti2019bert}
Chris Alberti, Kenton Lee, and Michael Collins. 2019.
\newblock A bert baseline for the natural questions.
\newblock \emph{arXiv preprint arXiv:1901.08634}.

\bibitem[{Camburu et~al.(2018)Camburu, Rockt{\"a}schel, Lukasiewicz, and
  Blunsom}]{camburu2018snli}
Oana-Maria Camburu, Tim Rockt{\"a}schel, Thomas Lukasiewicz, and Phil Blunsom.
  2018.
\newblock e-snli: Natural language inference with natural language
  explanations.
\newblock In \emph{Advances in Neural Information Processing Systems}, pages
  9539--9549.

\bibitem[{Camburu et~al.(2019)Camburu, Shillingford, Minervini, Lukasiewicz,
  and Blunsom}]{camburu2019make}
Oana-Maria Camburu, Brendan Shillingford, Pasquale Minervini, Thomas
  Lukasiewicz, and Phil Blunsom. 2019.
\newblock \href {http://arxiv.org/abs/1910.03065} {Make up your mind!
  adversarial generation of inconsistent natural language explanations}.

\bibitem[{Carlson(1977)}]{carlson1977unified}
Greg~N Carlson. 1977.
\newblock A unified analysis of the english bare plural.
\newblock \emph{Linguistics and philosophy}, 1(3):413--457.

\bibitem[{Clark et~al.(2019)Clark, Lee, Chang, Kwiatkowski, Collins, and
  Toutanova}]{clark2019boolq}
Christopher Clark, Kenton Lee, Ming-Wei Chang, Tom Kwiatkowski, Michael
  Collins, and Kristina Toutanova. 2019.
\newblock Boolq: Exploring the surprising difficulty of natural yes/no
  questions.
\newblock \emph{arXiv preprint arXiv:1905.10044}.

\bibitem[{Clark(1975)}]{clark1975bridging}
Herbert~H Clark. 1975.
\newblock Bridging.
\newblock In \emph{Theoretical issues in natural language processing}.

\bibitem[{Clark and Marshall(1981)}]{clark1981definite}
Herbert~H Clark and Catherine~R Marshall. 1981.
\newblock Definite knowledge and mutual knowledge.
\newblock \emph{Elements of Discourse Understanding}.

\bibitem[{Devlin et~al.(2019)Devlin, Chang, Lee, and
  Toutanova}]{Devlin2019BERTPO}
Jacob Devlin, Ming-Wei Chang, Kenton Lee, and Kristina Toutanova. 2019.
\newblock Bert: Pre-training of deep bidirectional transformers for language
  understanding.
\newblock \emph{ArXiv}, abs/1810.04805.

\bibitem[{Doshi-Velez and Kim(2017)}]{doshi2017towards}
Finale Doshi-Velez and Been Kim. 2017.
\newblock Towards a rigorous science of interpretable machine learning.
\newblock \emph{arXiv preprint arXiv:1702.08608}.

\bibitem[{Ehsan et~al.(2018)Ehsan, Harrison, Chan, and
  Riedl}]{ehsan2018rationalization}
Upol Ehsan, Brent Harrison, Larry Chan, and Mark~O Riedl. 2018.
\newblock Rationalization: A neural machine translation approach to generating
  natural language explanations.
\newblock In \emph{Proceedings of the 2018 AAAI/ACM Conference on AI, Ethics,
  and Society}, pages 81--87.

\bibitem[{Ehsan et~al.(2019)Ehsan, Tambwekar, Chan, Harrison, and
  Riedl}]{ehsan2019automated}
Upol Ehsan, Pradyumna Tambwekar, Larry Chan, Brent Harrison, and Mark~O Riedl.
  2019.
\newblock Automated rationale generation: a technique for explainable ai and
  its effects on human perceptions.
\newblock In \emph{Proceedings of the 24th International Conference on
  Intelligent User Interfaces}, pages 263--274. ACM.

\bibitem[{Goodrich et~al.(2020)Goodrich, Gabry, Ali, and Brilleman}]{rstanarm}
Ben Goodrich, Jonah Gabry, Imad Ali, and Sam Brilleman. 2020.
\newblock \href {https://mc-stan.org/rstanarm} {rstanarm: {Bayesian} applied
  regression modeling via {Stan}.}
\newblock R package version 2.19.3.

\bibitem[{Hearst(1992)}]{hearst1992automatic}
Marti~A Hearst. 1992.
\newblock Automatic acquisition of hyponyms from large text corpora.
\newblock In \emph{Proceedings of the 14th conference on Computational
  linguistics-Volume 2}, pages 539--545. Association for Computational
  Linguistics.

\bibitem[{Jacovi and Goldberg(2020)}]{jacovi2020towards}
Alon Jacovi and Yoav Goldberg. 2020.
\newblock Towards faithfully interpretable nlp systems: How should we define
  and evaluate faithfulness?
\newblock \emph{arXiv preprint arXiv:2004.03685}.

\bibitem[{Jain and Wallace(2019)}]{jain2019attention}
Sarthak Jain and Byron~C Wallace. 2019.
\newblock Attention is not explanation.
\newblock \emph{arXiv preprint arXiv:1902.10186}.

\bibitem[{Joshi et~al.(2019)Joshi, Chen, Liu, Weld, Zettlemoyer, and
  Levy}]{Joshi2019SpanBERTIP}
Mandar Joshi, Danqi Chen, Yinhan Liu, Daniel~S. Weld, Luke Zettlemoyer, and
  Omer Levy. 2019.
\newblock Spanbert: Improving pre-training by representing and predicting
  spans.
\newblock \emph{Transactions of the Association for Computational Linguistics},
  8:64--77.

\bibitem[{Krifka(2003)}]{krifka2003bare}
Manfred Krifka. 2003.
\newblock Bare nps: kind-referring, indefinites, both, or neither?
\newblock In \emph{Semantics and linguistic theory}, volume~13, pages 180--203.

\bibitem[{Kumar and Talukdar(2020)}]{kumar2020nile}
Sawan Kumar and Partha Talukdar. 2020.
\newblock \href {http://arxiv.org/abs/2005.12116} {Nile : Natural language
  inference with faithful natural language explanations}.

\bibitem[{Kwiatkowski et~al.(2019)Kwiatkowski, Palomaki, Redfield, Collins,
  Parikh, Alberti, Epstein, Polosukhin, Kelcey, Devlin, Lee, Toutanova, Jones,
  Chang, Dai, Uszkoreit, Le, and Petrov}]{47761}
Tom Kwiatkowski, Jennimaria Palomaki, Olivia Redfield, Michael Collins, Ankur
  Parikh, Chris Alberti, Danielle Epstein, Illia Polosukhin, Matthew Kelcey,
  Jacob Devlin, Kenton Lee, Kristina~N. Toutanova, Llion Jones, Ming-Wei Chang,
  Andrew Dai, Jakob Uszkoreit, Quoc Le, and Slav Petrov. 2019.
\newblock Natural questions: a benchmark for question answering research.
\newblock \emph{Transactions of the Association of Computational Linguistics}.

\bibitem[{Lamm et~al.(2018)Lamm, Chaganty, Manning, Jurafsky, and
  Liang}]{lamm2018textual}
Matthew Lamm, Arun~Tejasvi Chaganty, Christopher~D Manning, Dan Jurafsky, and
  Percy Liang. 2018.
\newblock Textual analogy parsing: Identifying what's shared and what's
  compared among analogous facts.
\newblock \emph{EMNLP}.

\bibitem[{Lee et~al.(2017)Lee, He, Lewis, and Zettlemoyer}]{Lee2017EndtoendNC}
Kenton Lee, Luheng He, Mike Lewis, and Luke Zettlemoyer. 2017.
\newblock End-to-end neural coreference resolution.
\newblock In \emph{EMNLP}.

\bibitem[{Lipton(2001)}]{lipton2001good}
Peter Lipton. 2001.
\newblock What good is an explanation?
\newblock In \emph{Explanation}, pages 43--59. Springer.

\bibitem[{Lipton(2016)}]{lipton2016mythos}
Zachary~C Lipton. 2016.
\newblock The mythos of model interpretability.
\newblock \emph{arXiv preprint arXiv:1606.03490}.

\bibitem[{Mikkelsen(2011)}]{mikkelsen2011copula}
Line Mikkelsen. 2011.
\newblock Copular clauses.
\newblock In Claudia Maienborn, Klaus von Heusinger, and Paul Portner, editors,
  \emph{Semantics: An International Handbook of Natural Language Meaning},
  volume~2, pages 1805--1829. Mouton De Gruyter, Berlin.

\bibitem[{Miltsakaki et~al.(2004)Miltsakaki, Prasad, Joshi, and
  Webber}]{miltsakaki2004penn}
Eleni Miltsakaki, Rashmi Prasad, Aravind~K Joshi, and Bonnie~L Webber. 2004.
\newblock The penn discourse treebank.
\newblock In \emph{LREC}.

\bibitem[{Min et~al.(2020)Min, Michael, Hajishirzi, and
  Zettlemoyer}]{min2020ambigqa}
Sewon Min, Julian Michael, Hannaneh Hajishirzi, and Luke Zettlemoyer. 2020.
\newblock \href {http://arxiv.org/abs/2004.10645} {Ambigqa: Answering ambiguous
  open-domain questions}.

\bibitem[{Mohankumar et~al.(2020)Mohankumar, Nema, Narasimhan, Khapra,
  Srinivasan, and Ravindran}]{mohankumar2020transparent}
Akash~Kumar Mohankumar, Preksha Nema, Sharan Narasimhan, Mitesh~M. Khapra,
  Balaji~Vasan Srinivasan, and Balaraman Ravindran. 2020.
\newblock \href {http://arxiv.org/abs/2004.14243} {Towards transparent and
  explainable attention models}.

\bibitem[{Narang et~al.(2020)Narang, Raffel, Lee, Roberts, Fiedel, and
  Malkan}]{narang2020wt5}
Sharan Narang, Colin Raffel, Katherine Lee, Adam Roberts, Noah Fiedel, and
  Karishma Malkan. 2020.
\newblock \href {http://arxiv.org/abs/2004.14546} {Wt5?! training text-to-text
  models to explain their predictions}.

\bibitem[{Pradhan et~al.(2012)Pradhan, Moschitti, Xue, Uryupina, and
  Zhang}]{Pradhan2012CoNLL2012ST}
Sameer Pradhan, Alessandro Moschitti, Nianwen Xue, Olga Uryupina, and Yuchen
  Zhang. 2012.
\newblock Conll-2012 shared task: Modeling multilingual unrestricted
  coreference in ontonotes.
\newblock In \emph{EMNLP-CoNLL Shared Task}.

\bibitem[{Rajpurkar et~al.(2016)Rajpurkar, Zhang, Lopyrev, and
  Liang}]{Rajpurkar_2016}
Pranav Rajpurkar, Jian Zhang, Konstantin Lopyrev, and Percy Liang. 2016.
\newblock \href {https://doi.org/10.18653/v1/d16-1264} {Squad: 100,000+
  questions for machine comprehension of text}.
\newblock \emph{Proceedings of the 2016 Conference on Empirical Methods in
  Natural Language Processing}.

\bibitem[{Reddy et~al.(2019)Reddy, Chen, and Manning}]{Reddy_2019}
Siva Reddy, Danqi Chen, and Christopher~D. Manning. 2019.
\newblock \href {https://doi.org/10.1162/tacl_a_00266} {Coqa: A conversational
  question answering challenge}.
\newblock \emph{Transactions of the Association for Computational Linguistics},
  7:249–266.

\bibitem[{Ross et~al.(2017)Ross, Hughes, and Doshi-Velez}]{ross2017right}
Andrew~Slavin Ross, Michael~C Hughes, and Finale Doshi-Velez. 2017.
\newblock Right for the right reasons: Training differentiable models by
  constraining their explanations.
\newblock \emph{arXiv preprint arXiv:1703.03717}.

\bibitem[{Russell(1905)}]{russell1905denoting}
Bertrand Russell. 1905.
\newblock On denoting.
\newblock \emph{Mind}, 14(56):479--493.

\bibitem[{Tandon et~al.(2019)Tandon, Dalvi, Sakaguchi, Clark, and
  Bosselut}]{tandon2019wiqa}
Niket Tandon, Bhavana Dalvi, Keisuke Sakaguchi, Peter Clark, and Antoine
  Bosselut. 2019.
\newblock Wiqa: A dataset for “what if...” reasoning over procedural text.
\newblock In \emph{Proceedings of the 2019 Conference on Empirical Methods in
  Natural Language Processing and the 9th International Joint Conference on
  Natural Language Processing (EMNLP-IJCNLP)}, pages 6078--6087.

\bibitem[{Tomasello et~al.(2007)Tomasello, Carpenter, and
  Liszkowski}]{tomasello2007new}
Michael Tomasello, Malinda Carpenter, and Ulf Liszkowski. 2007.
\newblock A new look at infant pointing.
\newblock \emph{Child development}, 78(3):705--722.

\bibitem[{Wiegreffe and Pinter(2019)}]{wiegreffe2019attention}
Sarah Wiegreffe and Yuval Pinter. 2019.
\newblock Attention is not not explanation.
\newblock \emph{arXiv preprint arXiv:1908.04626}.

\bibitem[{Wolfson et~al.(2020)Wolfson, Geva, Gupta, Gardner, Goldberg, Deutch,
  and Berant}]{Wolfson_2020}
Tomer Wolfson, Mor Geva, Ankit Gupta, Matt Gardner, Yoav Goldberg, Daniel
  Deutch, and Jonathan Berant. 2020.
\newblock \href {https://doi.org/10.1162/tacl_a_00309} {Break it down: A
  question understanding benchmark}.
\newblock \emph{Transactions of the Association for Computational Linguistics},
  8:183–198.

\end{thebibliography}
\bibliographystyle{acl_natbib}

\end{document}